\documentclass[10pt,twocolumn,letterpaper]{article}

\usepackage{iccv}
\usepackage{times}
\usepackage{epsfig}
\usepackage{graphicx}
\usepackage{amsmath}
\usepackage{amssymb}

\usepackage[pagebackref=true,breaklinks=true,letterpaper=true,colorlinks,bookmarks=false]{hyperref}

\iccvfinalcopy 

\def\iccvPaperID{2~AI-HADR} 

\ificcvfinal\fi

\begin{document}

\title{Disaster Anomaly Detector via Deeper FCDDs for Explainable Initial Responses}

\author{Takato Yasuno, Masahiro Okano and Junichiro Fujii\\
Research Institute for Infrastructure Paradigm Shift, Yachiyo Engineering,Co.,Ltd.\\
Taito-ku, Tokyo, 111-8648, Japan\\
{\tt\small \{tk-yasuno,ms-okano,jn-fujii\}@yachiyo-eng.co.jp}
}

\maketitle
\ificcvfinal\thispagestyle{empty}\fi

\begin{abstract}
Extreme natural disasters can have devastating effects on both urban and rural areas. 
In any disaster event, an initial response is the key to rescue within 72 hours and prompt recovery. 
During the initial stage of disaster response, it is important to quickly assess the damage over a wide area and identify priority areas. 
Among machine learning algorithms, deep anomaly detection is effective in detecting devastation features that are different from everyday features. 
In addition, explainable computer vision applications should justify the initial responses. 
In this paper, we propose an anomaly detection application utilizing deeper fully convolutional data descriptions (FCDDs), that enables the localization of devastation features and visualization of damage-marked heatmaps. 
More specifically, we show numerous training and test results for a dataset AIDER with the four disaster categories: collapsed buildings, traffic incidents, fires, and flooded areas. 
We also implement ablation studies of anomalous class imbalance and the data scale competing against the normal class.
Our experiments provide results of high accuracies over 95\% for $F_1$.
Furthermore, we found that the deeper FCDD with a VGG16 backbone consistently outperformed other baselines CNN27, ResNet101, and Inceptionv3. 
This study presents a new solution that offers a disaster anomaly detection application for initial responses with higher accuracy and devastation explainability, providing a novel contribution to the prompt disaster recovery problem in the research area of anomaly scene understanding.
Finally, we discuss future works to improve more robust, explainable applications for effective initial responses.
\end{abstract}

\section{Introduction}
\subsection{Related applications for disaster response}
The application of drones has rapidly progressed in a variety of fields over the past decade and is now being used in disaster management and humanitarian aid\cite{Mohd2022}. Mohd et al. summarized 52 research papers published between 2009 and 2020, outlining the trend of drone use in natural disasters, including landslides\cite{Chang2020}, hurricanes\cite{Martin2020}, and forest fires\cite{Tran2020}. These were classified into four categories: 1) mapping and disaster assessment, 2) search and rescue for victim identification, 3) emergency transportation, and 4) training and preparing emergency medical services (EMSs). 
First, drones are valuable for monitoring, mapping, and assessing initial damage. Drones are more effective than first responder information\cite{Ulfa2019}, as they can convey high-resolution images in less time than satellite imagery. In adittion, drones are capable of sharing aerial imagery and information more efficiently during the initial stages of a disaster.     
Second, the drone has frequently been used for search and rescue and can perform rescue operations in a variety of situations. Drones have also been used as a search tool to detect victims widely at sea\cite{Lygouras2019}.   
Third, previous studies\cite{Vaishali2019}\cite{Yakushiji2020} demonstrated the successful transport of disaster medical equipment, AEDs, insulin, and emergency food. Fourth, drones can be a resource for training and preparing EMS personnel to improve their ability to visualize aerial perspectives in disaster situations.
Therefore, the initial response to a disaster is crucial for increasing the capability of damage assessment and victim search and rescue preparations in humanitarian aid.

\subsection{Related models for deep anomaly detection}
Modern anomaly detection approaches can be divided into four categories: pixel-wise segmentation, one-class classification, patch-wise embedding similarity, and reconstruction-based models\cite{Yasuno2023}. Inspired by \cite{Ruff2020}, these anomaly detection approaches are reviewed in a unified manner, progressing from a lower to higher complexity scale through several categories of localization models. First, anomaly detection approaches based on a lower complexity scale include the one-class support vector machine (OC-SVM) \cite{Chalapathy2018}, support vector data description (SVDD) \cite{Tax2014}, principal component analysis (PCA) \cite{Hawkins1974}, and kernel-PCA \cite{Hoffmann2007}. Anomaly detection approaches based on a higher complexity scale include deep SVDD \cite{Ruff2018}, fully convolutional data description (FCDD) \cite{Liznerski2021}, variational autoencoders (VAEs) \cite{Kingma2019,An2015}, and adversarial auto-encoders (AAEs) \cite{Zhou2017}. However, owing to their susceptibility to background noise, reconstruction-based models are not always able to reconstruct synthetic outputs accurately. In contrast, one-class classification models depend on neither synthetic reconstruction nor probabilistic assumptions; therefore, they may be more robust anomaly detectors. 
However, the applicability of deep anomaly detection has not been well known in natural disaster situations. In particular, one-class classification models are expected to be an explainable application for robust disaster detection in complex background noise.  

Briefly, Rippel et al. \cite{Rippel2023,Yasuno2023} reviewed embedding-similarity based anomaly detection approaches in context of automated visual inspection (AVI). 
The patch-wise embedding approach enables to minimize the background noise per each patch image. To localize the anomalous feature, patch-wise embedding-similarity models perform that the normal reference can be the sphere feature containing embeddings from normal images. In this case, anomaly score is the distance between embedding vectors of a test image and reference vectors representing normality from the dataset. 
Embedding-similarity based models includes the SPADE \cite{Cohen2020}, PaDiM \cite{Thomas2020}, PatchCore \cite{Roth2021}, FastFlow \cite{Yu2021}. 
However, these models requires normalizing probability distribution and optimization algorithms such as a greedy coreset selection, a nearest neighbor search on a set of normal embedding vectors, so the inference complexity scales linearly to the size of training dataset. In contrast, one-class classification approach can learn efficiently using rare class of imbalanced dataset with fewer data scale in disaster damage detection for initial responses.

\subsection{Explainable detector for initial response}
For explainable disaster response in the initial stage, aerial image classifications have been proposed using unmanned aerial vehicles (UAVs)\cite{AIDER2019}, aerial photography\cite{asahi2020}. 
Kyrkou et al. provided transfer learning results with an average accuracy of 88.5–91.9\% using the VGG16, ResNet50, and MobileNet, and generated gradient-based heatmaps using the Grad-CAM\cite{Selvaraju2017}. Unfortunately, the accuracy of each disaster category was unknown. The supervised classification approach requires a large number of devastation images by category, and data mining is time-consuming because of the very low frequency of extreme events. The disadvantage of generating gradient-based heatmap is the need for parallel computing resources and iterative computation time.

The study highlights robustness of disaster scene detectior and explainability of devastation images. 
The authors think the initial response to a disaster is key for successful capability of assessing devastation damage, searching for victims and prioritizing rescue.
One-class classification models are expected to be an explainable application for robust disaster scene understanding behind the complex background.  
To explain damage-marked heatmaps, the direct upsampling approach potentially would help to reduce computational resource and running times. 
However, the aspects of robustness and explainability have not been sufficiently explored in the previous research on disaster detection. 
In this study, we propose a novel disaster detection application using fewer limited devastation images, and adopt the receptive field upsampling approach to explain disaster features for making initial response decisions.        

Section 2 outlines robust application methodology for explainable disaster detection. In section 3, applied results to disaster dataset and ablation studies on deeper backbones are demonstrated. Furthermore, quantitative comparisons to the previous models are summarized. Section 4 discusses the concluding remarks on explainable disaster detector application and future works to imbalanced devastation data.  

\section{Explainable Disaster Detection}
\subsection{One-class disaster classification}
The authors \cite{Yasuno2023} have already formulated the deeper FCDDs and found their applicability to damage data sets of bridges, dams, and buildings in the inspection for condition-based maintenance to make a decision to repair them. 
However, the deeper FCDDs has been not yet well known to feasible to disaster devastated images that includes inconsistent distance with different angle towards the damage location. As a hurdle to overcome, disaster images have unseen feature and complex background. 
For prompt disaster responses, visualizing devastated-mark heatmaps and computing hazard-weighted anomaly scores is critical for damage assessment and humanitarian assistance for effective recovery under scarce resources: time, medicine, medical team and equipments. 
For a novel appplication of explainable disaster detection, we summarize the application methodology of the deeper FCDDs, such as the robust loss function, hazard-weighted anomaly socres that are modified in this study, and the damage-marked heatmaps using the Gaussian upsampling algorithm. 

Let $D_i$ be the $i$-th disaster image with a size of $h\times w$. 
We consider the number of training images and the weight $W$ of the fully convolutional network (FCN). The deep support vector data description (SVDD) objective function \cite{Ruff2018} is formulated as a minimization problem for a deep SVDD.
A pseudo-Huber loss function is introduced to obtain a more robust loss formulation \cite{Ruff2021icml}. Let $\ell(z)$ be the loss function and define the pseudo-Huber loss function as follows:   
\begin{equation}
\ell(z) = \exp(-F(z)),~ F(z) = \sqrt{\|z\|^2 + 1} -1.
\end{equation}
Let $x_i =1$ denote the anomalous label of the $i$th disaster image, and let $x_i =0$ denote the normal label of the $i$th disaster image. 
Therefore, a loss function of our deeper FCDD with a deeper backbone $\nu$ can be formulated as follows.
\begin{equation}
\begin{split}
&\mathcal{L}_{deeperFCDD} = \frac{1}{n} \sum_{i=1}^{n} \frac{(1-x_i)}{uv} \sum_{x,y} F_{x,y} (\Xi^{\nu}_W(D_i)) \\ 
                           &- x_i \log \left[ 1 -  \exp\left\{ \frac{-1}{uv} \sum_{x,y} F_{x,y} (\Xi^{\nu}_W(D_i)) \right\} \right],
\end{split}
\end{equation}
where $F_{x,y}(u)$ are the elements $(x,y)$ of the receptive field of size $u\times v$ under a deeper FCDD. 
For initial disaster responses incorporating the hazard of death and injury, we propose a hazard-weight behind the disaster image for the initial stage of priority to search and rescue, effectively. 
The hazard-weighted anomaly score $H_i$ of the $i$th image is expressed as the sum of all the elements of the receptive field.
\begin{equation}
H_i(D_i~;~h_i,\nu) = h_i \sum_{x,y} F_{x,y} (\Xi^{\nu}_W(D_i)),~i=1,\cdots,n.
\end{equation}
were, $h_i$ is the weight of the hazard devastated by a natural disaster. For example, in densely populated regions, the weight of disaster hazards can be set higher. 
Specifically, we provided the $i$th ratio of the curve to match the GNSS-based position.   
In this paper, present the construction of a baseline FCDD \cite{Yasuno2023} with an initial backbone $\nu=0$ and performed CNN27 mapping $\Xi^0_W(D_i)$ from the input image $D_i$ in the dataset. We also present deeper FCDDs focusing on deeper backbones $\nu\in \{$VGG16, ResNet101, Inceptionv3$\}$ with a mapping operation $\Xi^{\nu}_W(D_i)$ to achieve a more accurate detection.  
We conducted ablation studies on class imbalance and data scale applied to a disaster dataset to build an anomaly detection application for initial responses. 
\subsection{Explainable damage-mark heatmap}
Convolutional neural network (CNN) architectures, comprising millions of parameters, have exhibited remarkable performance for damage detection. Heatmap visualization techniques for localizing anomalous features are typically categorized as masked sampling \cite{Zeiler2013}\cite{Ribeiro2016}, class activation maps (CAMs)\cite{Zhou2015} approaches, and gradient-based extensions (Grad-CAM) \cite{Selvaraju2017}. The disadvantages of the aforementioned methods are their requirement for parallel computation resources and iterative computation time. 
This would lead the waste of time and crucial delay of starts for damage mapping, search, and rescue. 

In this study for explainable initial response applications, we adopt the receptive field upsampling approach \cite{Liznerski2021} to emphasize devastation features using an upsampling-based activation map with Gaussian upsampling from the receptive field. The primary advantages of the upsampling approach are the reduced computational resource requirements and shorter computation times. The Gaussian upsampling algorithm generates a full-resolution heatmap from the input of a low-resolution receptive field $u\times v$.
 
Let $F\in {R}^{u\times v}$ be a low-resolution receptive field (input), and let $F'\in {R}^{h\times w}$ be a full resolution of the hazard-mark heatmap (output).
We define the 2D Gaussian distribution $G_2(a_1,a_2,\delta)$ as follows: 
\begin{equation}
\begin{split}
&[G_2(a_1,a_2,\delta)]_{x,y} \equiv \\
 &\frac{1}{2\pi\delta^2}\exp\left(-\frac{(x-a_1)^2+(y-a_2)^2}{2\delta^2}\right).  
\end{split}
\end{equation}
The Gaussian upsampling algorithm from the receptive field is implemented as follows:
\begin{enumerate}
\item $F' \leftarrow 0 \in {R}^{h\times w}$
\item for all output pixels $q$ in $F \leftarrow 0 \in {R}^{u\times v}$
\item \qquad $u(q) \leftarrow$ is upsampled from a receptive field of $q$
\item \qquad $(c_1(u),c_2(u)) \leftarrow$ is the center of the field $u(q)$
\item \qquad $F' \leftarrow F' + q\cdot G_2(c_1,c_2,\delta)$
\item end for
\item return $F'$ 
\end{enumerate}
After conducting experiments with various datasets, we determined that a receptive field size of $28 \times 28$ is a practical value. When generating a hazardous heatmap, unlike a revealed damage mark, we need to unify the display range corresponding to the anomaly scores, ranging from the minimum to the maximum value. To strengthen the defective regions and highlight the hazard marks, we define a display range of [min, max/4], where the quartile parameter is 0.25. This results in the histogram of anomaly scores having a long-tailed shape. If we were to include the complete anomaly score range, the colors would weaken to blue or yellow on the maximum side.

In this study, we utilized the Automated Visual Inspection Library by MathWorks Computer Vision Toolbox Team, released on MATLAB2023a. We implemented a single 8GB GPU equipped by the ELSA Quadro RTX4000.  

\section{Applied Results}
\subsection{Single class anomaly detection}
\subsubsection{Datasets of natural disaster}
There are several datasets of natural disaster devastation by hurricanes \cite{texas2018}, typhoons \cite{asahi2020}, earthquakes \cite{sakurada2015}, and combinations of multiple disasters, including collapsed buildings, traffic incidents, fires, and floods \cite{AIDER2019}.
These disaster images were collected using various modes, including satellite imagery, aerial photography, drone-based systems, and panoramic 360-degree cameras. 
In this study, we used the open-access AIDER \cite{AIDER2019} dataset with four categories collected as aerial drone images for the capability of damage assessment and preparing search and rescue at the initial response.
Table~\ref{tab:dataAIDER} shows four disaster classes and the number of images that are classified as normal and anomalous. Note that the number of fires class is 518, because we deleted 3 images that were expressed as POND5.    

\begin{table}[h]
  \begin{center}
\begin{tabular}{|c|c|r|r}
\hline
Disaster class & Normal & Anomalous \\
\hline\hline
collapsed buildings  & 2000 & 511 \\
traffic incidents & 2000 & 485 \\
fires & 2000 & 518 \\
flooded areas & 2000 & 526 \\
\hline
\end{tabular}
  \end{center}
\caption{\label{tab:dataAIDER}Disaster dataset using drones \cite{AIDER2019} with four classes and the number of images.}
\end{table}

\subsubsection{Training results and accuracy}
During the training of the anomaly detector, we fixed the input size to $224^2$. To train the model, we set the mini-batch size to 32 and ran 50 epochs. We used the Adam optimizer with a learning rate of 0.0001, a gradient decay factor of 0.9, and a squared gradient decay factor of 0.99. The training images were partitioned at a ratio of 65:15:20 for training, calibration, and testing.
As shown in Table~\ref{tab:accBuild},~\ref{tab:accTraffic},~\ref{tab:accFire},~\ref{tab:accFlood}, our deeper FCDD based on VGG16 (FCDD-VGG16) outperformed the baseline and other backbone-based deeper FCDDs in the disaster dataset ~\ref{tab:dataAIDER} for detecting anomalies of each class.
For an initial stage of disaster damage assessment, the recall is critical to minimize the error of false positive that can not discriminate anomalous feature. When we predicted test images using our deeper FCDD-VGG16, all of recall scores had over 97\% rather than the existing average accuracy 91.9\% using the state-of-the-art of classification model applied to the AIDER dataset \cite{AIDER2019}.   

\begin{table}[h]
  \begin{center}
\begin{tabular}{|c|c|c|c|c|}
\hline
Backbone & AUC & $F_1$ & Precision & Recall \\
\hline
\hline
CNN27 & 0.9623 & 0.8803 & 0.8598 & 0.9019 \\
\textbf{VGG16} & \textbf{0.9979} & \textbf{0.9611} & \textbf{0.9519} & \textbf{0.9705} \\
ResNet101 &0.9943 & 0.9604 & 0.9700 & 0.9509 \\
Inceptionv3 &0.9940 & 0.9423 & 0.9245 & 0.9607 \\
\hline
\end{tabular}
  \end{center}
\caption{\label{tab:accBuild}Collapsed buildings single class anomaly detection: backbone ablation studies using deeper FCDDs.}
\end{table}

\begin{table}[h]
  \begin{center}
\begin{tabular}{|c|c|c|c|c|}
\hline
Backbone & AUC & $F_1$ & Precision & Recall \\
\hline
\hline
CNN27 & 0.9504 & 0.7907 & 0.7203 & 0.8762 \\
\textbf{VGG16} & \textbf{0.9769} & \textbf{0.9178} & \textbf{0.8636} & \textbf{0.9793} \\
ResNet101 &0.9719 & 0.8942 & 0.8378 & 0.9587 \\
Inceptionv3 &0.9708 & 0.9108 & 0.8761 & 0.9484 \\
\hline
\end{tabular}
  \end{center}
\caption{\label{tab:accTraffic}Traffic incidents single class anomaly detection: backbone ablation studies using deeper FCDDs.}
\end{table}

\begin{table}[h]
  \begin{center}
\begin{tabular}{|c|c|c|c|c|}
\hline
Backbone & AUC & $F_1$ & Precision & Recall \\
\hline
\hline
CNN27 & 0.9547 & 0.8558 & 0.8288 & 0.8846 \\
\textbf{VGG16} & \textbf{0.9991} & \textbf{0.9760} & \textbf{0.9714} & \textbf{0.9807} \\
ResNet101 &0.9989 & 0.9668 & 0.9532 & 0.9807 \\
Inceptionv3 &0.9967 & 0.9528 & 0.9351 & 0.9711 \\
\hline
\end{tabular}
  \end{center}
\caption{\label{tab:accFire}Fires single class anomaly detection: backbone ablation studies using deeper FCDDs.}
\end{table}

\begin{table}[h]
  \begin{center}
\begin{tabular}{|c|c|c|c|c|}
\hline
Backbone & AUC & $F_1$ & Precision & Recall \\
\hline
\hline
CNN27 & 0.9784 & 0.7949 & 0.7089 & 0.9047 \\
\textbf{VGG16} & \textbf{0.9984} & \textbf{0.9589} & \textbf{0.9210} & \textbf{1.000} \\
ResNet101 &0.9924 & 0.9581 & 0.9363 & 0.9809 \\
Inceptionv3 &0.9887 & 0.9519 & 0.9611 & 0.9428 \\
\hline
\end{tabular}
  \end{center}
\caption{\label{tab:accFlood}Flooded areas single class anomaly detection: backbone ablation studies using deeper FCDDs.}
\end{table}

\subsubsection{Disaster damage-mark heatmaps}
\begin{figure}[h]
\centering
\includegraphics[width=0.38\textwidth]{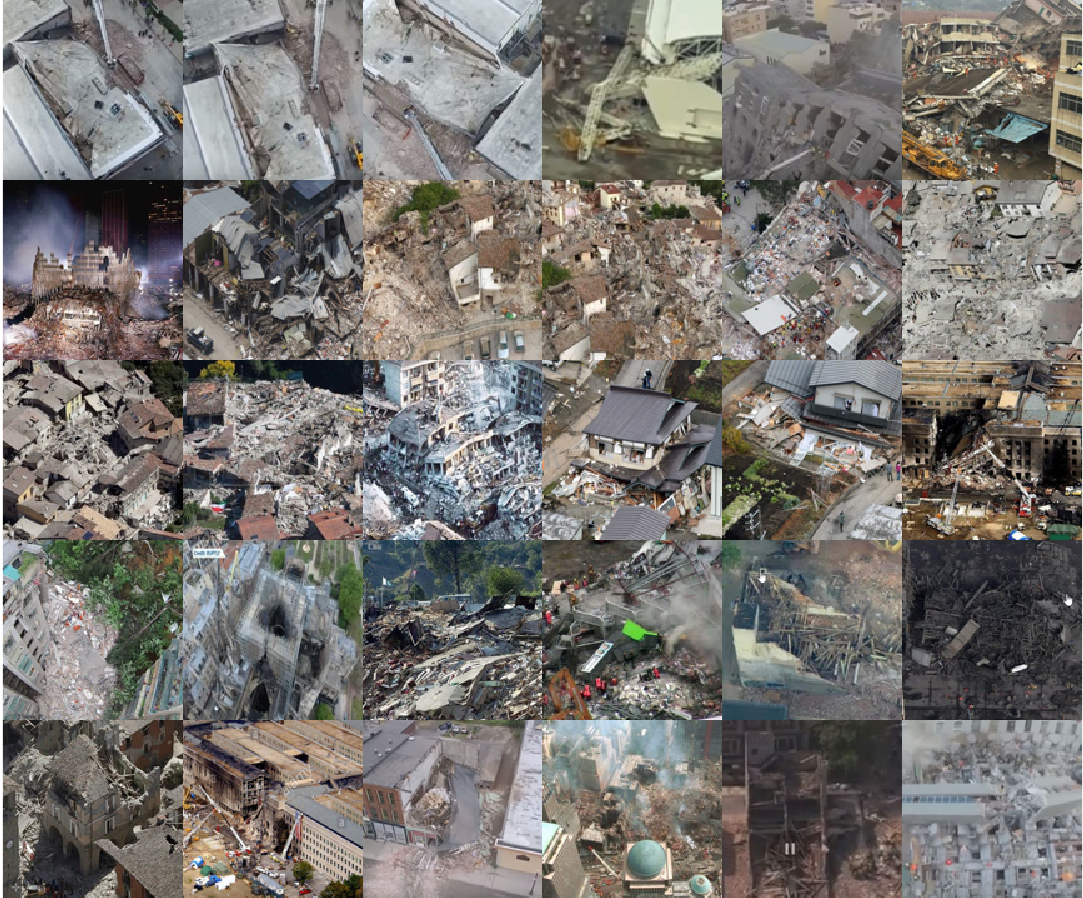} \\
\includegraphics[width=0.38\textwidth]{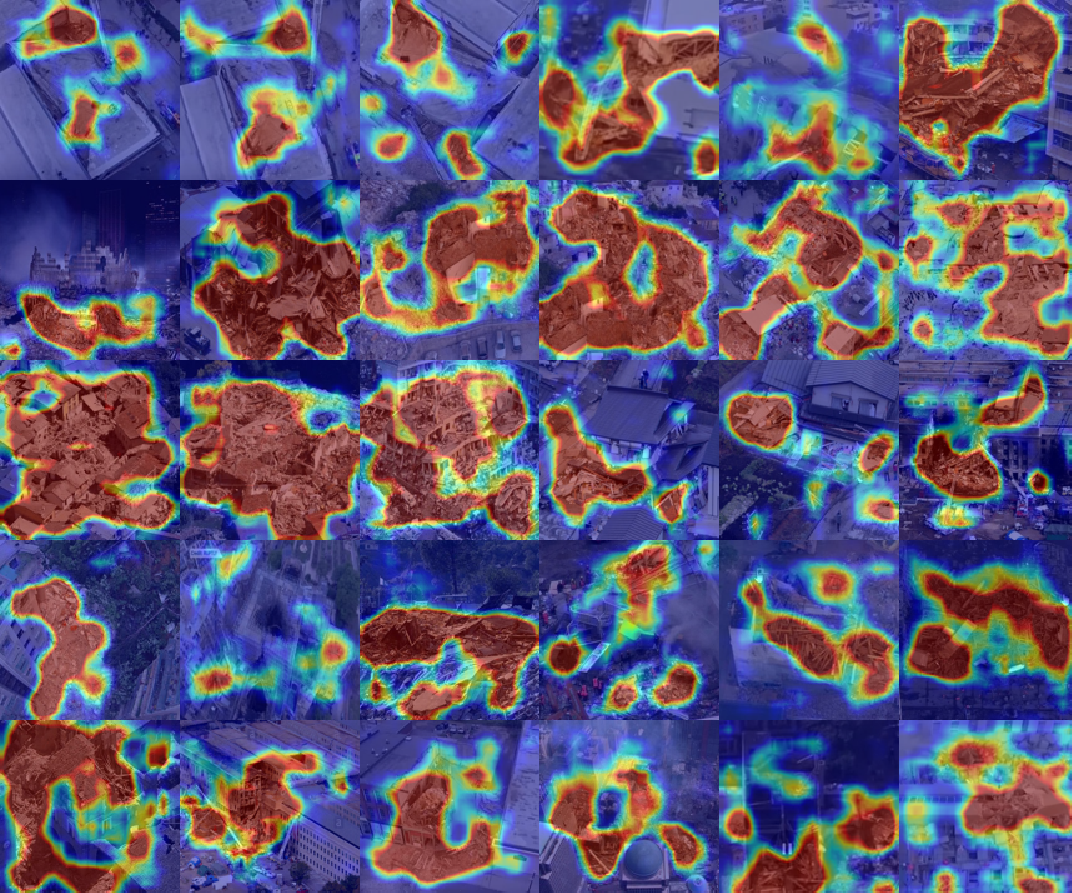}
\caption{\label{fig:rawBuild}Collapsed buildings: raw images (top) and damage mark heatmaps (bottom) using the VGG16 backbone.}
\end{figure}

\begin{figure}[h]
\centering
\includegraphics[width=0.37\textwidth]{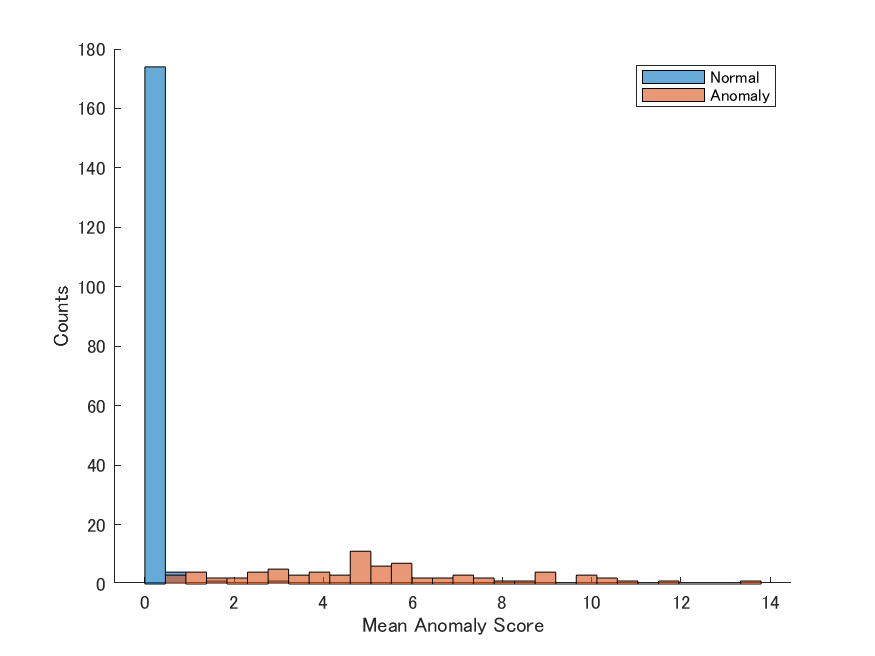}
\caption{\label{fig:histBuild}Histogram of collapsed buildings scores corresponding to our deeper FCDD on the VGG16 backbone.}
\end{figure}

We visualized each devastation feature by Gaussian upsampling in the deeper FCDD with a promising VGG16 backbone, and generated a histogram of the anomaly scores of the test images for each disaster class. 
In Figure \ref{fig:rawBuild}, the red region represents the collapsed buildings of devastation feature, that are almost fully covered on heatmaps using our Gaussian upsampling.
We express the red region as the {\it damage-mark}. This damage-marked heatmaps suprisingly emphasize the devastation feature rather than the previous gradient-based heatmaps \cite{AIDER2019}. 
Because the maximum of display range are set the qurtile 0.25 according to the long-tailed shape of the histogram, as shown in Figure \ref{fig:histBuild}. This illustrates that one overlapping bin exists in the horizontal anomaly scores. For detecting collapsed buildings, the score range was well separated.

\begin{figure}[h]
\centering
\includegraphics[width=0.38\textwidth]{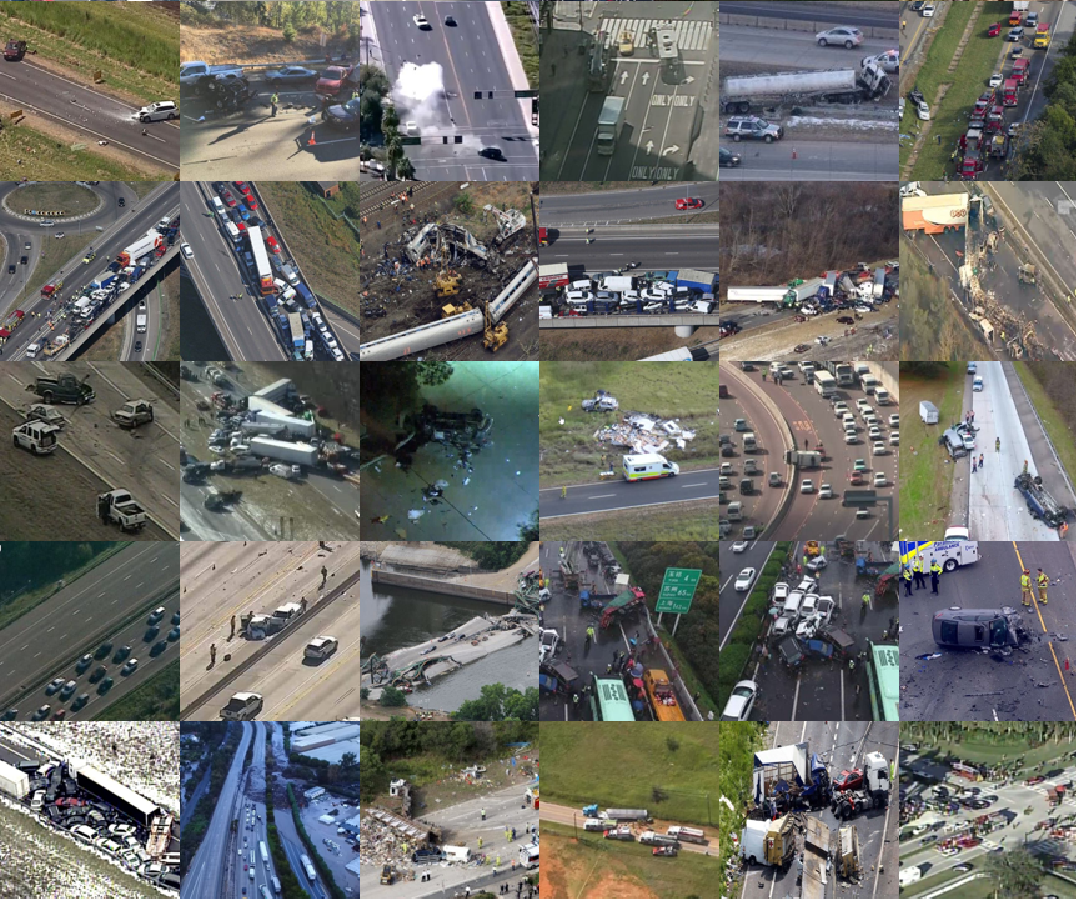} \\
\includegraphics[width=0.38\textwidth]{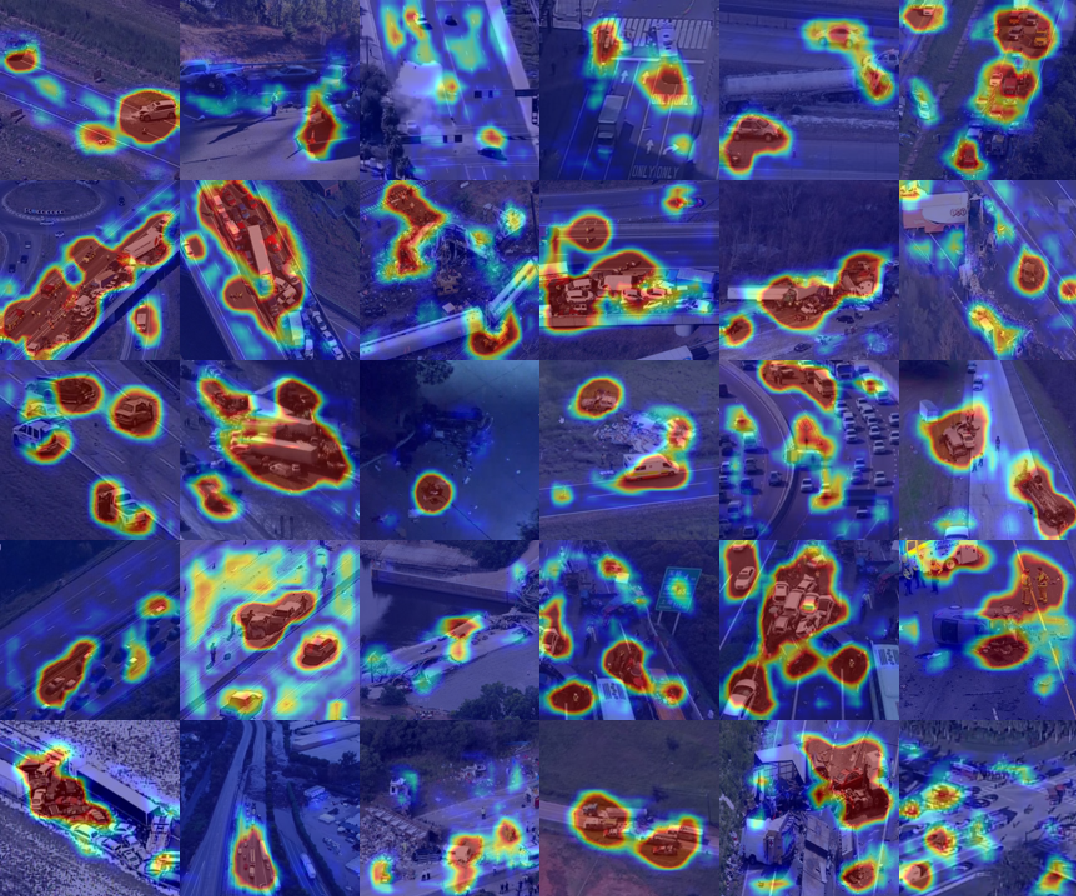}
\caption{\label{fig:rawTraff}Traffic incidents: raw images (top) and damage mark heatmaps (bottom) using the VGG16 backbone.}
\end{figure}

\begin{figure}[h]
\centering
\includegraphics[width=0.37\textwidth]{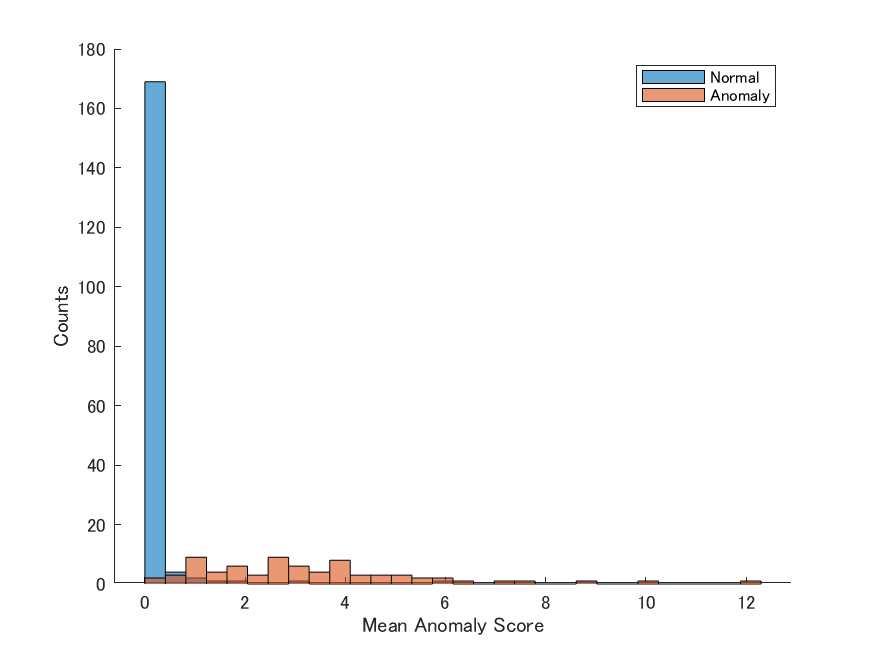}
\caption{\label{fig:histTraff}Histogram of traffic incidents scores corresponding to our deeper FCDD on the VGG16 backbone.}
\end{figure}

\begin{figure}[h]
\centering
\includegraphics[width=0.38\textwidth]{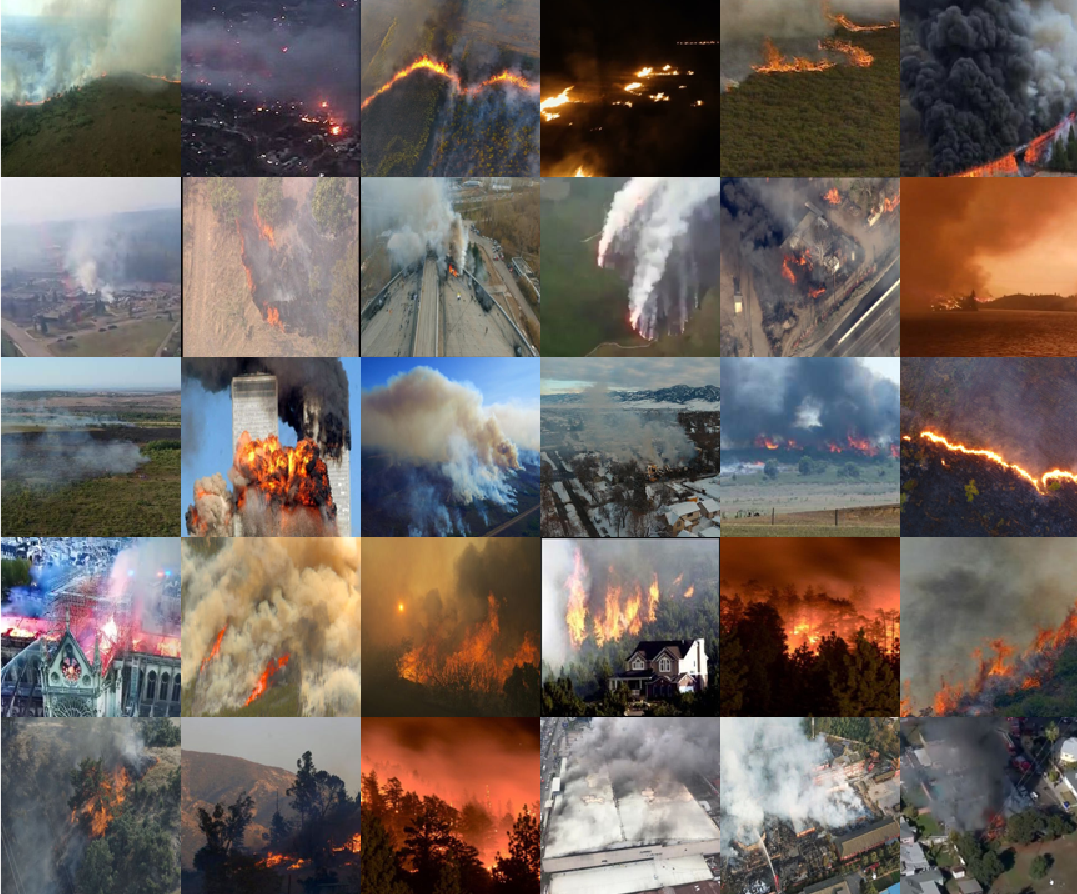} \\
\includegraphics[width=0.38\textwidth]{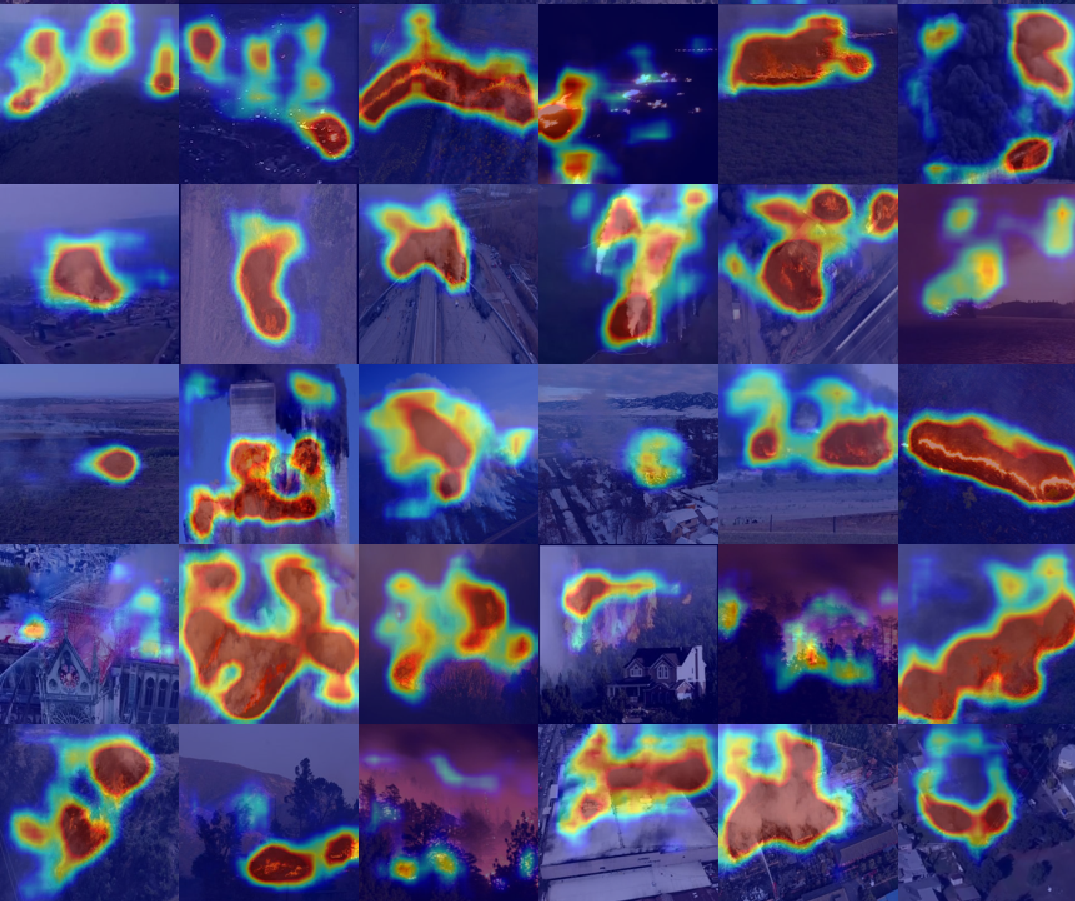}
\caption{\label{fig:rawFire}Fires: raw images (top) and damage mark heatmaps (bottom)  using the VGG16 backbone.}
\end{figure}

\begin{figure}[h]
\centering
\includegraphics[width=0.37\textwidth]{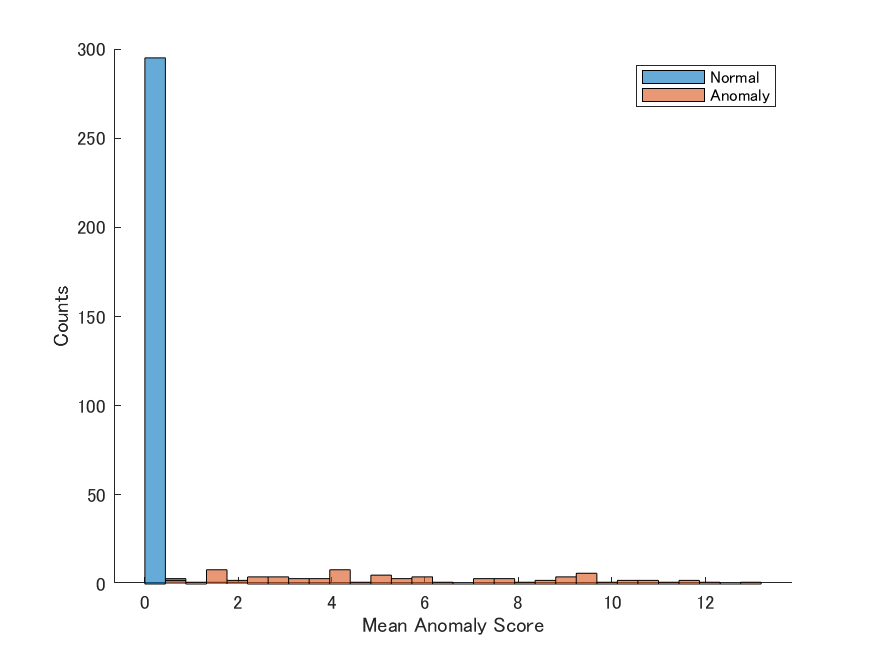}
\caption{\label{fig:histFire}Histogram of fires scores corresponding to our deeper FCDD on the VGG16 backbone.}
\end{figure}

\begin{figure}[h]
\centering
\includegraphics[width=0.38\textwidth]{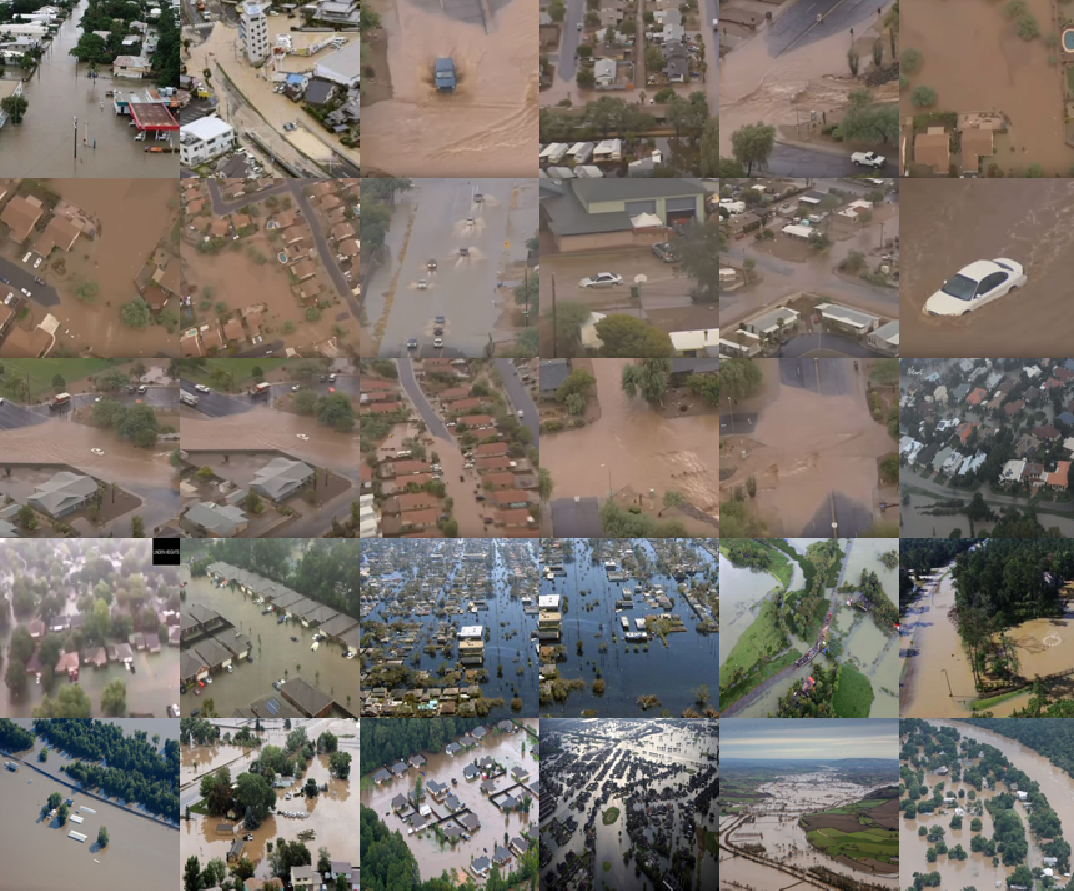} \\
\includegraphics[width=0.38\textwidth]{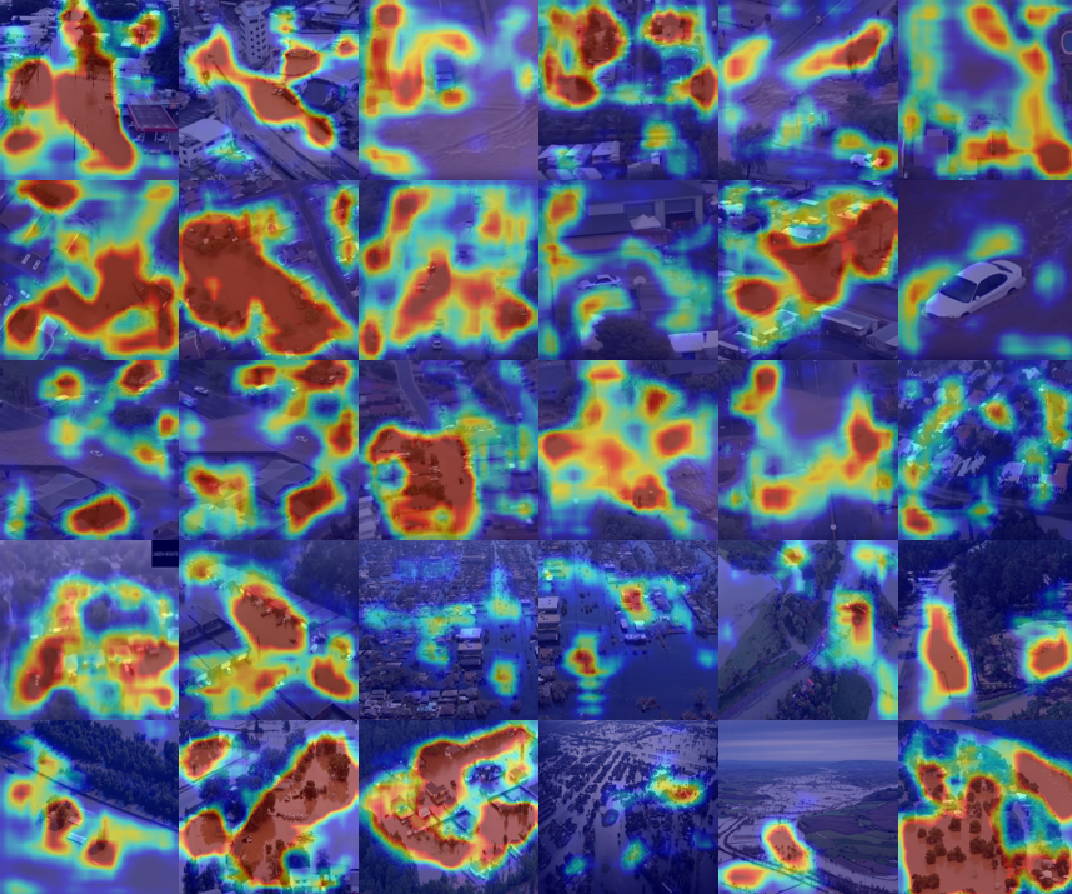}
\caption{\label{fig:rawFlood}Flooded areas: raw images (top) and damage mark heatmaps (bottom) using the VGG16 backbone.}
\end{figure}

\begin{figure}[h]
\centering
\includegraphics[width=0.37\textwidth]{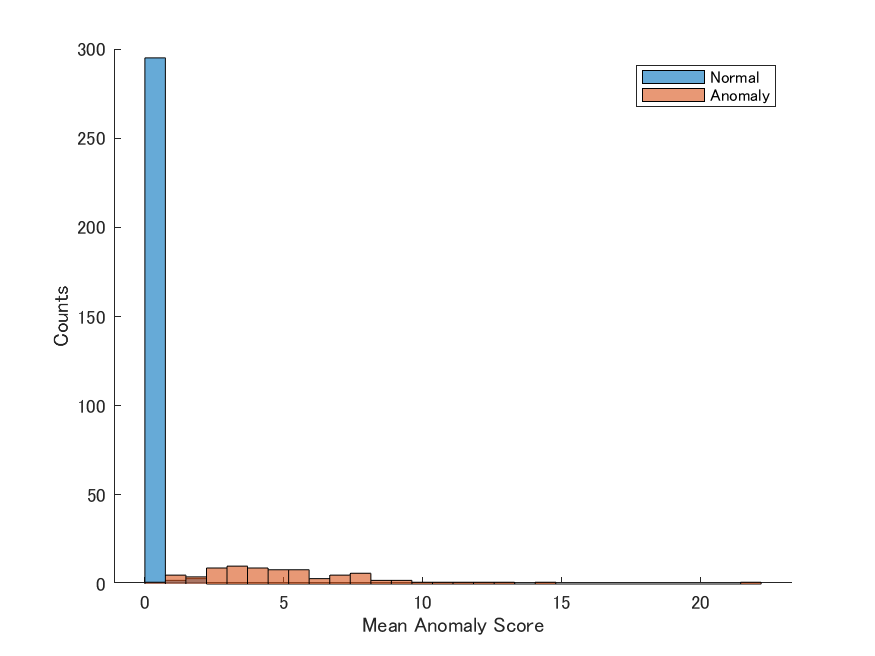}
\caption{\label{fig:histFlood}Histogram of flood areas scores corresponding to our deeper FCDD on the VGG16 backbone.}
\end{figure}

As shonw in Figure \ref{fig:rawTraff}, the damage-marks enhance to the traffic incident feature on heatmaps, in which represents that cars/trains are overturned, fallen sideways, extremely congested in each incident scene. 
Figure \ref{fig:histTraff} illustrates that several overlapping bins exist in the horizontal anomaly scores. For detecting traffic incidents, the score range was well separated. 
In Figure \ref{fig:rawFire}, the damage-marks emphasize the fires and smokes. Figure \ref{fig:histFire} illustrates that a few overlapping bins exist in the horizontal anomaly scores. For detecting fires and smokes, the score range is well separated.
Figure \ref{fig:rawFlood} depicts the flooded areas of damage-mark emphasizing the surface of muddy river on heatmaps. Figure \ref{fig:histFlood} shows that a few overlapping bins remain in the anomaly scores. For detecting flooded areas, the score range was well separated. 

Thus, we found that the {\it damage-mark} represented the devastation feature, that were almost fully covered on our Gaussian upsampling heatmaps.
Furthermore, this damage-marked heatmaps suprisingly emphasize the devastation feature rather than the previous Grad-CAM heatmaps \cite{AIDER2019}.




\subsection{Pooled classes anomaly detection}
\subsubsection{Datasets for ablations on imbalance and scale}
Table~\ref{tab:dataPool} shows the dataset that pooled four disaster categories for ablation studies of class imbalance and data scale. Here, one thousand anomalous images contain each class of 250 randomly sampled from the original set, respectively. 
For imbalance studies, the number of normal classes changes every 1,000 skips from 1,000 to 4,000 towards the anomalous class of 1,000 images. 
In addition, for data scaling studies, the number of normal classes is set within three ranges from 2,000 to 4,000 towards the anomalous class of 2,000 images.

\begin{table}[h]
  \begin{center}
\begin{tabular}{|c|c|r|r|}
\hline
Dataset & Normal & Anomalous \\
\hline\hline
balanced    1K~:~1K & 1000 & 1000 \\
imbalanced 2K~:~1K & 2000 & 1000 \\
imbalanced 3K~:~1K & 3000 & 1000 \\
imbalanced 4K~:~1K & 4000 & 1000 \\
\hline
scaled \& balanced 2K~:~2K & 2000 & 2000 \\
scaled \& imbalanced 3K~:~2K & 3000 & 2000 \\
scaled \& imbalanced 4K~:~2K & 4000 & 2000 \\
\hline
\end{tabular}
  \end{center}
\caption{\label{tab:dataPool}Dataset of four pooled classes for ablation studies of class imbalance and data scale. Herein, anomalous 1K images contain each class of 250 randomly sampled from the original images.}
\end{table}

\subsubsection{Pooled training results and accuracy}
As presented in the upper four rows of Table~\ref{tab:ablationImbScale}, we highlighted imbalances, and implemented ablation studies using our deeper FCDD with a VGG16 backbone. Because of the limited anomalous images, less than 250 by category, the highest accuracy value is inconsistent and unstable. 
In contrast, in the lower three rows of Table~\ref{tab:ablationImbScale}, we added the data scaling, and addressed ablation studies using deeper FCDD-VGG16.    
In the case of 3K~,~2K, the $F_1$, and precision values are the highest in this study.
In terms of AUC and recall, the case 4K~,~2K is the highest value.  
Thus, we found that the anomalous class requires over 2,000 images and the normal class scales more than 3,000 images for the pooled disaster dataset.

For simultaneous multi-class of damage assessment for initial responses, both of precision and recall, i.e. $F_1$, are crucial to minimize the false negative error and false positive error. In other words, explainable disaster detection should recognize the devastation pattern, accurately and thoroughly. 
When we predicted the pooled test images using our deeper FCDD-VGG16, all of $F_1$ scores performed over 95\% rather than the existing average accuracy 91.9\% using the previous classification model applied to the AIDER dataset \cite{AIDER2019}. 

\begin{table}[h]
  \begin{center}
\begin{tabular}{|c|c|c|c|c|}
\hline
norm. , anom. & AUC & $F_1$ & Precision & Recall \\
\hline
\hline
1K~,~1K & 0.9819 & 0.9343 & 0.9438 & 0.9250 \\
2K~,~1K & 0.9882 & 0.9181 & 0.9113 & 0.9250 \\
3K~,~1K & 0.9917 & 0.9090 & 0.8515 & 0.9750 \\
4K~,~1K & 0.9949 & 0.9326 & 0.8981 & 0.9700 \\
\hline
2K~,~2K  &0.9886 & 0.9497 & 0.9545 & 0.9450 \\
\textbf{3K~,~2K } & 0.9946 & \textbf{0.9527} & \textbf{0.9713} & 0.9348 \\
\textbf{4K~,~2K } & \textbf{0.9974} & 0.9507 & 0.9591 & \textbf{0.9423} \\
\hline
\end{tabular}
  \end{center}
\caption{\label{tab:ablationImbScale}Pooled 4 classes anomaly detection: ablation studies on imbalance and the scale using our deeper FCDD on the VGG16. Norm. and anom. denotes normal and anomalous, respectively.}
\end{table}

\subsubsection{Pooled disaster damage-mark heatmaps}
\begin{figure}[h]
\centering
\includegraphics[width=0.38\textwidth]{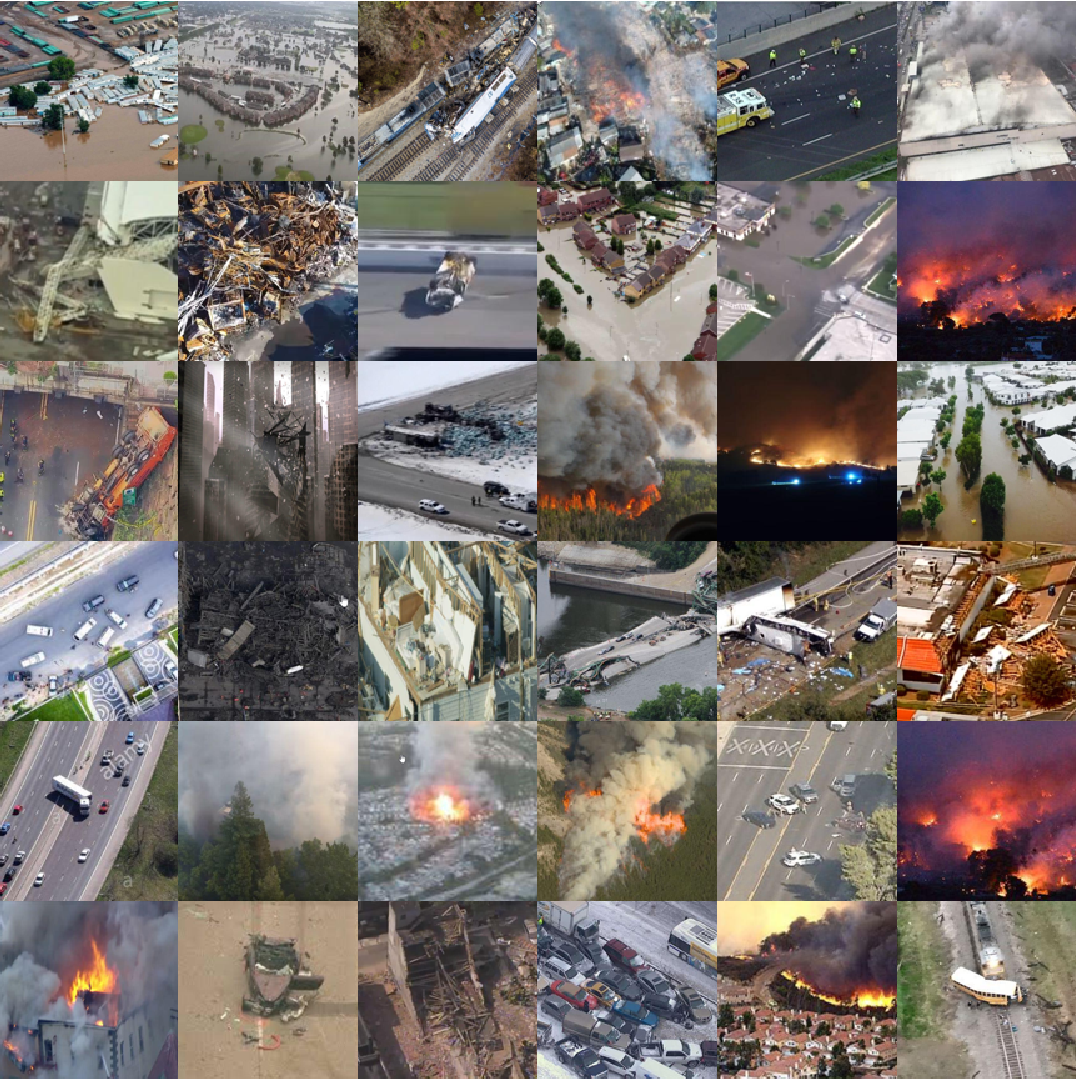} \\
\includegraphics[width=0.38\textwidth]{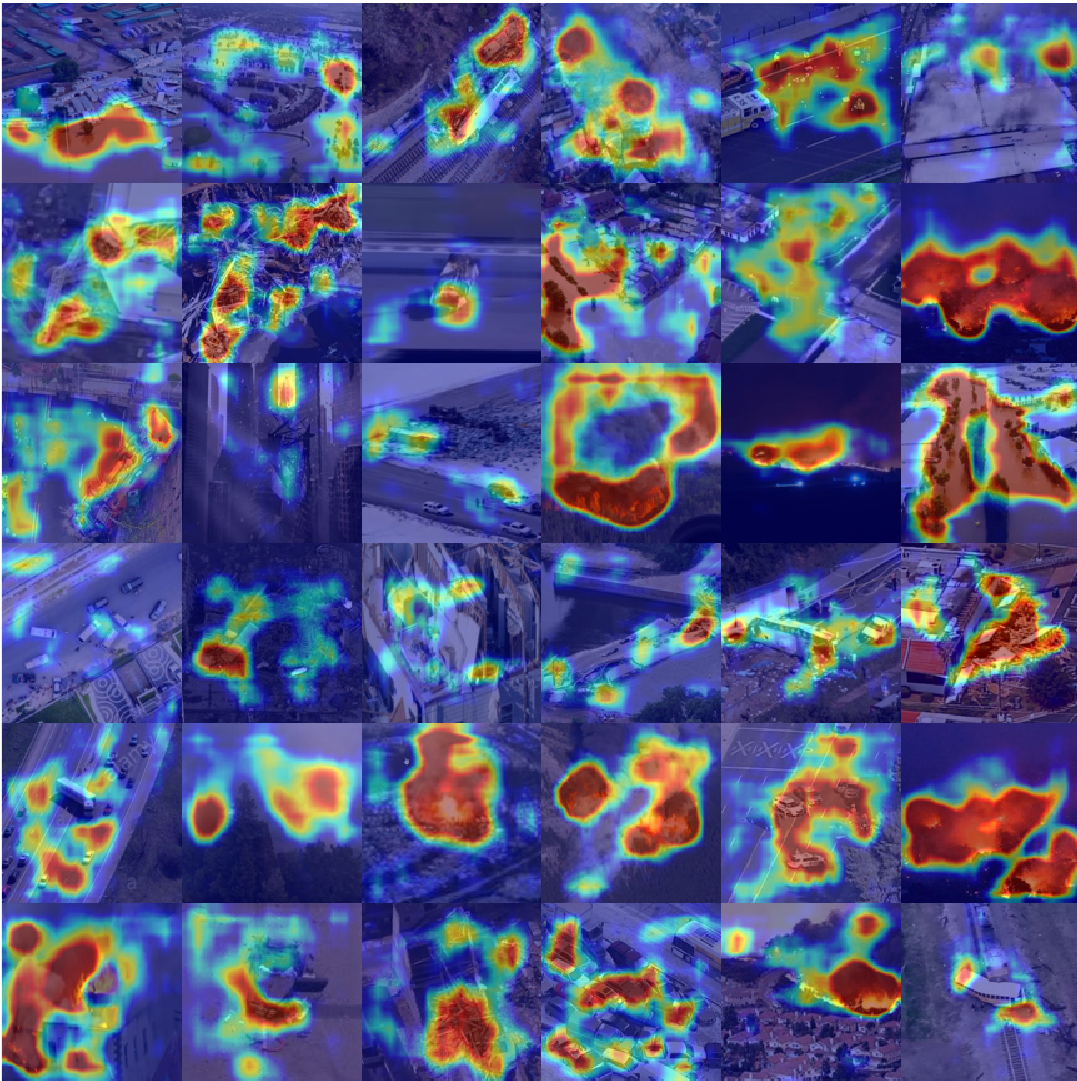}
\caption{\label{fig:rawPool}Pooled four classes that randomly sampled three thousand normal and two thousand pooled anomalies: raw images (top) and damage mark heatmaps (bottom) using our deeper FCDD on VGG16 backbone.}
\end{figure}

\begin{figure}[h]
\centering
\includegraphics[width=0.37\textwidth]{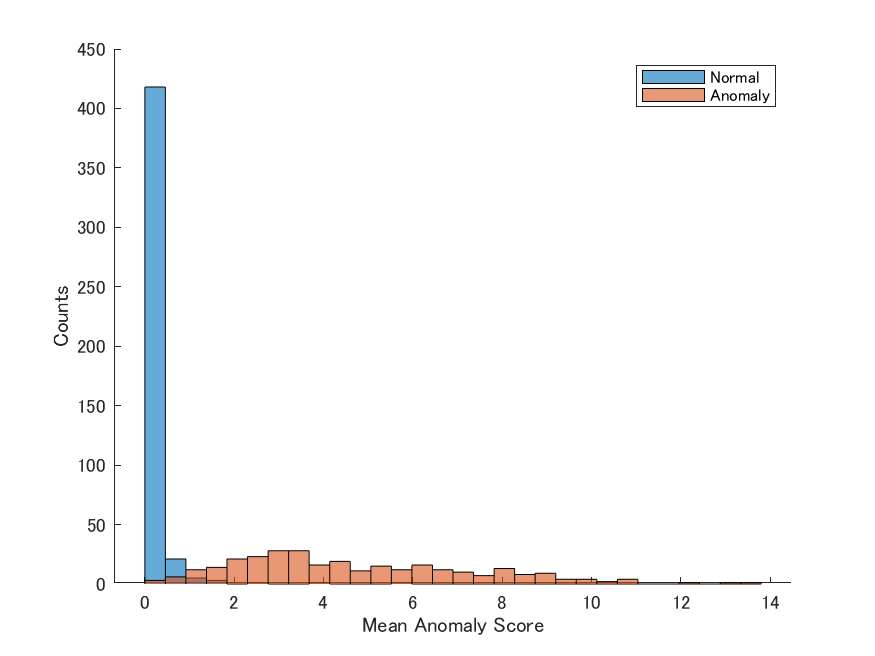}
\caption{\label{fig:histPool}Histogram of pooled four classes scores corresponding to our deeper FCDD on VGG16 backbone.}
\end{figure}

We visualized the disaster features by using Gaussian upsampling, and generated a histogram of the anomaly scores of the test images for the pooled disaster dataset. 
Figure \ref{fig:rawPool} presents {\it damage-mark} heatmaps using deeper FCDD-VGG16. The damage-marks emphasize the devastation regions of multi-class of categories: collapsed buildings, traffic incidents, fires and smoke, and flooded areas. 
Figure \ref{fig:histPool} illustrates that a few overlapping bins exist in the horizontal anomaly scores. For detecting multi-class of disaster categories, the score range was well separated.

Therefore, we found that the {\it damage-mark} represented multi-class of devastation region, that were almost fully covered on our Gaussian upsampling heatmaps.
Significantly, this damage-marked heatmaps emphasize multi-class of devastation feature rather than the previous gradient-based heatmaps \cite{AIDER2019}.


\begin{table*}[h]
  \begin{center}
\begin{tabular}{|c|c|c|c|c|c|c|c|c|}
\hline
Accuracy & VAE & PaDiM-R18 & PaDiM-R50 & PatchCore-R18 & PatchCore-R50 & FCDD-CNN27 & \textbf{deeperFCDD} \\
\hline
\hline
AUC        & --     & 0.7450 & 0.7962 & 0.7308 & 0.7812 & 0.8957 & \textbf{0.9819} \\
$F_1$      &0.6373 & 0.7014 & 0.7317 & 0.6868 & 0.7333 & 0.7989 & \textbf{0.9343} \\
Precision &0.6060 & 0.5826 & 0.6459 & 0.6045 & 0.6032 & 0.8360 & \textbf{0.9438} \\
Recall      &0.6720 & 0.8812 & 0.8437 & 0.7950 & 0.9350 & 0.7650 & \textbf{0.9250} \\
\hline
\end{tabular}
  \end{center}
\caption{\label{tab:comparison}Accuracy comparisons of the previous deep anomaly detection models, applied to the dataset of four pooled classes whose number of normal and anomalous images are 1000, respectively. The R18 denotes the backbone of ResNet18, and the R50 indicates the one of ResNet50. The deeperFCDD has the promissing backbone of VGG16 that outperformed than others in our ablation studies.}
\end{table*}

\subsection{Quantitative comparisons}
As summarized in Table~\ref{tab:comparison}, we quantitatively compared our deeper FCDDs to the previous deep anomaly detection models. For their accuracy comparisons, we implemented experimental studies to the disaster dataset pooled with 4 categories. The normal and anomalous images are randomly sampled 1,000 respectively, just the same as the balanced case 1K~:~1K in Table~\ref{tab:dataPool}.
As the one of reconstruction-based models \cite{Kingma2019,An2015}, the VAE performed worse accuracy than the PaDiM \cite{Thomas2020}) on the ResNet50. Because the normal scene has many background noise, the synthetic normal image by auto-encoding is incompletely reconstructed for computing the recunstruction loss. 
Moreover, in terms of $F_1$, the PaDiMs have worse accuracy than the core set sub-sampling model, the PatchCore \cite{Roth2021} on the ResNet50. 
Because of the normal scene has complex feature such as sky, grass, various angle of forward vision, the embedding similarity-based learner may not sufficiently recognize the pattern of core feature in normal images by drones.   

The baseline FCDD \cite{Liznerski2021} using the shallow CNN27 architecture, achieved higher accuracy than the patch-wise embedding similarity-based models in the field of disaster scene understanding.
Furthermore, our classification-based anomaly detection application, deeper FCDD using the deeper backbone of the VGG16, outperformed than other previous embedding similarity-based models. 
Thus, we found that our deeper FCDD-VGG16 promised to be robust to disaster scene and high performance of devastation detection for initial responses. 

\section{Concluding Remarks}
\subsection{Explainable disaster detection application}
This study proposed an anomaly detection application utilizing our deeper FCDDs that enables localization of devastation features and visualization of damage-marked heatmaps. 
We provided our deeper FCDDs trained results and damage-marks emphasized heatmaps to an open-access disaster dataset with the four categories: collapsed buildings, traffic incidents, fires, and flooded areas. 
Our experiments provided promising results with higher accuracies around 91–97\% for $F_1$. In each disaster category, deeper FCDDs created the damage-marks for visual explanation, even without annotating the devastation regions.
Furthermore, we implemented ablation studies of class imbalance from 1~:~1 to 4~:~1,
 and the data scaling ranged from two to four thousand, where the highest accuracies were over 95\% for $F_1$.
In this study, we found that the deeper FCDD with a VGG16 backbone (FCDD-VGG16) consistently outperformed other baselines CNN27, ResNet101, and Inceptionv3. 
This study presents a new solution that offers a disaster anomaly detection application for initial responses with higher accuracy and devastation explainability, providing a novel contribution to the field of disaster scene recognition for prompt response.
\subsection{Future work on limited data and imbalance}
We discuss future work to improve more robust, explainable applications for effective initial responses.
This study investigated the feasibility of disaster anomaly detection by highlighting four categories of natural disasters. However, this scope is too limited for disaster recovery decision-making and prompt responses. We find it difficult to detect all disasters using the anomaly detection application, therefore, we need to target other disasters such as earthquake, tsunami, rockfall, volcano, and heavy snowfall for more robust application.
To address the challenges of anomalous class imbalanced data, mixture augmentation preprocessing could be effective for classification-based models. However, the imbalance issue remains for the infrequent occurrences of natural disasters.
We intend to create natural disaster foundation models using data-scaled training and supervised devastation feature learning. 
To overcome the hurdle of narrow data space and scene locality, we should propose a methodology of domain adaptation based on a foundation model for disaster detection anywhere.  

\subsection*{Acknowledgment}
The authors wish to thank the MathWorks Computer Vision Toolbox Team for providing MATLAB helpful Labrary for Automated Visual Inspection and resouces for deep anomaly detection. 
{\small
\bibliographystyle{ieee_fullname}
\bibliography{egbib}

\begin{thebibliography}{10}\itemsep=-1pt

\bibitem{An2015}
Jinwon An and Sungzoon Cho.
\newblock Variational autoencoder based anomaly detection using reconstruction
  probability.
\newblock {\em Special Lecture on IE}, 2(1), 2015.

\bibitem{texas2018}
Quoc~Dung Cao and Youngjun Choe.
\newblock Building damage annotation on post-hurricane satellite imagery based
  on convolutional neural networks.
\newblock {\em Natural Hazards}, 103(3):3357--3376, jul 2020.

\bibitem{Chalapathy2018}
Raghavendra Chalapathy, Aditya~Krishna Menon, and Sanjay Chawla.
\newblock Anomaly detection using one-class neural networks, 2019.

\bibitem{Chang2020}
Kuo-Jen Chang, Chun-Wei Tseng, Chih-Ming Tseng, Ta-Chun Liao, and Ci-Jian Yang.
\newblock Application of unmanned aerial vehicle ({UAV})-acquired topography
  for quantifying typhoon-driven landslide volume and its potential topographic
  impact on rivers in mountainous catchments.
\newblock {\em Applied Sciences}, 10(17), 2020.

\bibitem{Cohen2020}
Niv Cohen and Yedid Hoshen.
\newblock Sub-image anomaly detection with deep pyramid correspondences, 2021.

\bibitem{Thomas2020}
Thomas Defard, Aleksandr Setkov, Angelique Loesch, and Romaric Audigier.
\newblock {PaDiM}: A patch distribution modeling framework for anomaly
  detection and localization.
\newblock In {\em Pattern Recognition. ICPR International Workshops and
  Challenges: Virtual Event, January 10–15, 2021, Proceedings, Part IV}, page
  475–489, Berlin, Heidelberg, 2021.

\bibitem{Hawkins1974}
{Douglas M.} Hawkins.
\newblock Detection of errors in multivariate data using principal components.
\newblock {\em Journal of the American Statistical Association},
  69(346):340--344, 1974.

\bibitem{Hoffmann2007}
Heiko Hoffmann.
\newblock Kernel pca for novelty detection.
\newblock {\em Pattern Recogn.}, 40(3):863–874, mar 2007.

\bibitem{Kingma2019}
Diederik~P. Kingma and Max Welling.
\newblock An introduction to variational autoencoders.
\newblock {\em Found. Trends Mach. Learn.}, 12(4):307–392, nov 2019.

\bibitem{AIDER2019}
C. Kyrkou and Theocharis Theocharides.
\newblock Deep-learning-based aerial image classification for emergency
  response applications using unmanned aerial vehicles.
\newblock {\em IEEE/CVF Conference on Computer Vision and Pattern Recognition
  Workshops ({CVPRW})}, pages 517--525, 2019.

\bibitem{Liznerski2021}
Philipp Liznerski, Lukas Ruff, Robert~A. Vandermeulen, Billy~Joe Franks, and
  Marius~Kloft et al.
\newblock Explainable deep one-class classification.
\newblock In {\em The International Conference on Learning
  Representations({ICLR})}, Workshop on Uncertainty and Robustness in Deep
  Learning, 2021.

\bibitem{Lygouras2019}
Eleftherios Lygouras, Nicholas Santavas, Anastasios Taitzoglou, Konstantinos
  Tarchanidis, Athanasios Mitropoulos, and Antonios Gasteratos.
\newblock Unsupervised human detection with an embedded vision system on a
  fully autonomous uav for search and rescue operations.
\newblock {\em Sensors}, 19(16), 2019.

\bibitem{Mohd2022}
Sharifah Mastura~Syed {Mohd Daud}, Mohd Yusmiaidil~Putera {Mohd Yusof},
  Chong~Chin Heo, Lay~See Khoo, Mansharan~Kaur {Chainchel Singh}, and Mohd
  Shah~Mahmood et al.
\newblock Applications of drone in disaster management: A scoping review.
\newblock {\em Science and Justice}, 62(1):30--42, 2022.

\bibitem{Vaishali2019}
Vaishali Nimilan, Gunaselvi Manohar, R.Sudha, and Pearley Stanley.
\newblock Drone-aid: An aerial medical assistance.
\newblock {\em International Journal of Innovative Technology and Exploring
  Engineering}, 8(11S), 2019.

\bibitem{Ribeiro2016}
Marco~Tulio Ribeiro, Sameer Singh, and Carlos Guestrin.
\newblock "why should i trust you?": Explaining the predictions of any
  classifier.
\newblock In {\em Proceedings of the 22nd {ACM SIGKDD} International Conference
  on Knowledge Discovery and Data Mining}, KDD '16, page 1135–1144.
  Association for Computing Machinery, 2016.

\bibitem{Rippel2023}
Oliver Rippel and Dorit Merhof.
\newblock Anomaly detection for automated visual inspection: A review.
\newblock In Volker Lohweg, editor, {\em Bildverarbeitung in der Automation},
  pages 1--13, Berlin, Heidelberg, 2023. Springer Berlin Heidelberg.

\bibitem{Roth2021}
Karsten Roth, Latha Pemula, Joaquin Zepeda, Bernhard Schölkopf, Thomas Brox,
  and Peter Gehler.
\newblock Towards total recall in industrial anomaly detection.
\newblock In {\em CVF Conference on Computer Vision and Pattern Recognition},
  page 14318–14328, 2021.

\bibitem{Ruff2020}
Lukas Ruff, Jacob~R. Kauffmann, Robert~A. Vandermeulen, Gregoire Montavon,
  Wojciech Samek, Marius Kloft, Thomas~G. Dietterich, and Klaus-Robert Muller.
\newblock A unifying review of deep and shallow anomaly detection.
\newblock {\em Proceedings of the {IEEE}}, 109(5):756--795, may 2021.

\bibitem{Ruff2018}
Lukas Ruff, Robert Vandermeulen, Nico Goernitz, Lucas Deecke, Shoaib~Ahmed
  Siddiqui, Alexander Binder, Emmanuel M{\"u}ller, and Marius Kloft.
\newblock Deep one-class classification.
\newblock In Jennifer Dy and Andreas Krause, editors, {\em Proceedings of the
  35th International Conference on Machine Learning}, volume~80 of {\em
  Proceedings of Machine Learning Research}, pages 4393--4402. PMLR, 10--15 Jul
  2018.

\bibitem{Ruff2021icml}
Lukas Ruff, Robert~A. Vandermeulen, Billy~Joe Franks, Klaus-Robert Müller, and
  Marius Kloft.
\newblock Rethinking assumptions in deep anomaly detection.
\newblock In {\em The International Conference on Machine Learning ({ICML})},
  Workshop on Uncertainty and Robustness in Deep Learning, 2021.

\bibitem{sakurada2015}
Ken Sakurada and Takayuki Okatani.
\newblock Change detection from a street image pair using cnn features and
  superpixel segmentation.
\newblock In {\em British Machine Vision Conference}, 2015.

\bibitem{Martin2020}
Martin Schaefer, Richard Teeuw, Simon Day, Dimitrios Zekkos, Paul Weber, and
  Toby~Meredith et al.
\newblock Low-cost {UAV} surveys of hurricane damage in dominica: automated
  processing with co-registration of pre-hurricane imagery for change analysis.
\newblock {\em Natural Hazards}, 101:755–784, 2020.

\bibitem{Selvaraju2017}
Ramprasaath~R. Selvaraju, Michael Cogswell, Abhishek Das, Ramakrishna Vedantam,
  and Devi et~al. Parikh.
\newblock {Grad-CAM}: Visual explanations from deep networks via gradient-based
  localization.
\newblock In {\em 2017 {IEEE} International Conference on Computer Vision
  ({ICCV})}, pages 618--626, 2017.

\bibitem{Tax2014}
David M.~J. Tax and Robert~P. Duin.
\newblock Support vector data description.
\newblock {\em Machine Learning}, 54(1):45--66, 2004.

\bibitem{Tran2020}
Dai~Quoc Tran, Minsoo Park, Daekyo Jung, and Seunghee Park.
\newblock Damage-map estimation using {UAV} images and deep learning algorithms
  for disaster management system.
\newblock {\em Remote Sensing}, 12(24), 2020.

\bibitem{Ulfa2019}
F. Ulfa and J. Sartohadi.
\newblock The role of small format aerial photographs for first response in
  landslide event.
\newblock {\em IOP Conference Series: Earth and Environmental Science},
  338(1):012026, nov 2019.

\bibitem{Yakushiji2020}
Koki Yakushiji, Hiroshi Fujita, Mikio Murata, Naoki Hiroi, Yuuichi Hamabe, and
  Fumiatsu Yakushiji.
\newblock Short-range transportation using unmanned aerial vehicles ({UAVs})
  during disasters in japan.
\newblock {\em Drones}, 4(4), 2020.

\bibitem{asahi2020}
Takato Yasuno, Masazumi Amakata, and Masahiro Okano.
\newblock Natural disaster classification using aerial photography explainable
  for typhoon damaged feature.
\newblock In {\em The 25th International Conference on Pattern
  Recognition({ICPR})}, Workshop on Machine Learning Advances Environmental
  Science({MAES}), 2020.

\bibitem{Yasuno2023}
Takato Yasuno, Masahiro Okano, and Junichiro Fujii.
\newblock One-class damage detector using deeper fully convolutional data
  descriptions for civil application.
\newblock {\em Advances in Artificial Intelligence and Machine Learning},
  3(2):996--1011, 2023.

\bibitem{Yu2021}
Jiawei Yu, Ye Zheng, Xiang Wang, Wei Li, Yushuang Wu, Rui Zhao, and Liwei Wu.
\newblock {FastFlow}: Unsupervised anomaly detection and localization via 2d
  normalizing flows, 2021.

\bibitem{Zeiler2013}
Matthew~D Zeiler and Rob Fergus.
\newblock Visualizing and understanding convolutional networks, 2013.

\bibitem{Zhou2015}
Bolei Zhou, Aditya Khosla, Agata Lapedriza, Aude Oliva, and Antonio Torralba.
\newblock Learning deep features for discriminative localization, 2015.

\bibitem{Zhou2017}
Chong Zhou and Randy~C. Paffenroth.
\newblock Anomaly detection with robust deep autoencoders.
\newblock In {\em Proceedings of 23rd ACM SIGKDD International Conference on
  Knowledge Discovery and Data Mining}, KDD '17, page 665–674, New York, USA,
  2017. Association for Computing Machinery.

\end{thebibliography}
}

\end{document}